\documentclass{article}
\usepackage{amsmath,amssymb,amsthm,cases,graphicx,subcaption}
\usepackage[utf8]{inputenc}
\usepackage[margin=1in]{geometry}
\usepackage{algorithm}

\usepackage{multirow}
\usepackage{bm}
\usepackage{microtype}

\usepackage{booktabs} 
\usepackage{natbib}
\setcitestyle{open={(},close={)}}

\usepackage[colorlinks=true,citecolor=blue]{hyperref}

\newtheorem{theorem}{Theorem}[section]

\usepackage{float}
\usepackage{epsfig,graphicx,verbatim}
\usepackage{pifont}% http://ctan.org/pkg/pifont

\newcommand{\markcir}{\ding{164}}
\newcommand{\marksqure}{\ding{110}}

\usepackage{tikz}
\usetikzlibrary{fit,shapes.geometric}

\newcounter{nodemarkers}
\newcommand\circletext[1]{%
    \tikz[overlay,remember picture] 
        \node (marker-\arabic{nodemarkers}-a) at (0,1.5ex) {};%
    #1%
    \tikz[overlay,remember picture]
        \node (marker-\arabic{nodemarkers}-b) at (0.,-0.02){};%
    \tikz[overlay,remember picture,inner sep=1pt]
        \node[draw,rectangle,fit=(marker-\arabic{nodemarkers}-a.center) (marker-\arabic{nodemarkers}-b.center)] {};%
    \stepcounter{nodemarkers}%
}

\begin{document}

\title{\textbf{SpaceE: Knowledge Graph Embedding\\ by Relational Linear Transformation in the Entity Space}}

\author{\vspace{0.5in}\\\textbf{Jinxing Yu, Yunfeng Cai, Mingming Sun, Ping Li}\vspace{0.2in}\\
Cognitive Computing Lab\\
 Baidu Research\\
 No. 10 Xibeiwang East Road, Beijing 10193, China\\
 10900 NE 8th St. Bellevue, Washington 98004, USA\\\\
 \{yujinxing,\ caiyunfeng,\ sunmingming01,\  liping11\}@baidu.com\\\\
}

\date{}

\maketitle

\begin{abstract}  

\vspace{0.2in}

\noindent\footnote{This work was initially submitted in 2019, to the ACL'20 conference. The authors sincerely thank the anonymous reviewers of ACL'20 and the reviewers of other conferences for our later submissions after ACL'20.}Translation distance based knowledge graph embedding (KGE) methods, such as \textit{TransE} and \textit{RotatE}, model the relation in knowledge graphs as translation or rotation in the vector space. Both translation and rotation are injective; that is, the translation or rotation of different vectors results in different results. In knowledge graphs, different entities may have a relation with the same entity; for example, many actors starred in one movie. Such a non-injective relation pattern cannot be well modeled by the translation or rotation operations in existing translation distance based KGE methods.  To tackle the challenge, we propose a translation distance-based KGE method called \textbf{SpaceE} to model relations as linear transformations. The proposed SpaceE embeds both entities and relations in knowledge graphs as matrices and  SpaceE  naturally models non-injective relations with singular linear transformations. We theoretically demonstrate that SpaceE is a fully expressive model with the ability to infer multiple desired relation patterns, including symmetry, skew-symmetry, inversion, Abelian composition, and non-Abelian composition. Experimental results on link prediction datasets illustrate that SpaceE substantially outperforms many previous translation distance based knowledge graph embedding methods, especially on datasets with many non-injective relations. The code is available based on the PaddlePaddle deep learning platform \url{https://www.paddlepaddle.org.cn/}.
\end{abstract}

\newpage

\section{Introduction} 
\vspace{0.1in}

Extracting entities and relationships from web texts~\citep{BankoOIE2007, sun2018logician, Proc:Sun_EMNLP18,NerSurvey2018, Dai2019CoreferenceAR, Smirnova_re_survey,  Proc:Liu_WWW20, Proc:Sun_EMNLP20,zhang2020Learning,zhang_readre,wang2022oie} and reasoning using the extracted information~\citep{moon-etal-2019-opendialkg, ChenKGReasoningReview2020} is the long-term goal of data-mining and natural language processing. From the philosophy of representation learning,  entities and relations (either natural texts or symbols) can be represented by embedding representations, and the reasoning can be performed by applying the mathematical operations on these embedding vectors. This methodology has been successfully applied in the field of knowledge graph embedding in both symbolic knowledge-bases~\citep{TransE, sun2018rotate, MQuadE} and open-domain textual fact corpus~\citep{gupta-etal-2019-care}, for the applications of knowledge graph completion~\citep{nguyen-2020-survey}, question answering~\citep{HuangKGEQA2019}, topic modeling~\citep{li2019Integration} and so on. 

The core theoretical problems of this methodology are to find appropriate the embedding form and the mathematical operation for reasoning. These problems have been mostly studied in the field of knowledge graph embedding for the task of knowledge graph completion. The background of knowledge graph completion is that: although the large scale knowledge graphs such as Yago~\citep{yago} and Freebase~\citep{freebase2008} store vast amounts of fact triples about the relationships between entities, knowledge graphs are incomplete and have missing relationships between entities. As the facts in knowledge graphs can be represented by $(head, relation, tail)$, the task of knowledge graph completion is to answer two queries: $(head, relation, ?)$ and $(?, relation, tail)$. Knowledge graph embedding (KGE) methods learn embedding representations of the entities and relations and use distance score functions to measure the plausibility of candidate fact triples. There are three major types of KGE methods: Translation distance-based methods model the relations as geometric transformations from the head entity vector to the tail entity vector and use norms as the score functions; Bilinear semantic matching methods model the plausibility of fact triples by bilinear semantic matching functions; Deep learning methods use deep neural networks to model the interaction between the head entity, the relation, and the tail entity of a fact triple.

The intuition behind the effectiveness of various design choices in KGE methods is to enable reasoning over the knowledge graph and to infer many relation patterns, including symmetry, skew-symmetry, inversion, and composition~\citep{sun2018rotate}. Furthermore, non-injective relations are prevalent in many knowledge graphs. For example, 98\% of the relations in FB15k-237 dataset are non-injective (1-N, N-1, N-N). Modeling the non-injective relations is challenging, especially when the N-side of the relation is large, because the KGE model has to handle the uncertainty in the knowledge graph completion. 

Among the translation distance based KGE methods, RotatE~\citep{sun2018rotate} is an expressive model. Although it achieves excellent performances, the non-injective relations are still challenging to model for RotatE. RotatE embeds the entities and relations as complex vectors, $\bm{h},  \bm{r},  \bm{t} \in \mathbb{C}^k, \|r_i\| = 1$. It is expected that $\bm{t} = \bm{h} \circ \bm{r}$, where $\circ$ denotes Hadamard product (elementwise). For two fact triples $(h, r, t_1), (h, r, t_2)$ of a 1-N relation $r$, the embeddings of different tail entities $t_1, t_2$ tend to be the same, that is $\bm{t}_1 = \bm{h} \circ \bm{r} = \bm{t}_2$. The reason for this undesired tendency is that rotation in complex vector space is injective.

BoxE~\citep{BoxE} is a recently proposed state-of-the-art translation distance based KGE method. It represents relations as a set of hyper-rectangles (or boxes). It has the ability to model non-injective relations well by using the box representation. Nonetheless, it cannot explicitly model the compositions of relations~\citep{BoxE}.

\vspace{0.1in}

\textbf{Contributions.} \ In this paper, we present a translation distance based knowledge graph embedding method called SpaceE that models relations as linear transformations in the entity space. Our contributions are summarized as follows: 1) We propose SpaceE, a novel knowledge graph embedding method to better model non-injective relations with non-injective linear transformations; 2) We theoretically demonstrate that SpaceE can infer relation patterns, including symmetry, skew-symmetry, inversion, Abelian composition, and non-Abelian composition; 3) We conduct extensive experiments on benchmarks and achieve comparable or better performances than previous translation distance based knowledge graph embedding methods. Experimental results demonstrate our model's superiority in modeling non-injective relations and its capability to infer various relation patterns.

\section{Related Work}
The task of link prediction in knowledge graphs has been extensively studied in the literature, and many methods have been proposed. Traditional approaches use rule-based logics~\citep{markovLogic}, or collect path features and use logistic regression on the features~\citep{PRA1, PRA2} for link prediction. Knowledge graph embedding methods later become popular given their simplicity, scalability, and better performance than traditional approaches. A few recent studies~\citep{mln_tang, mln_gcn} combine the rule-based Markov Logic Networks (MLNs) and knowledge graph embedding methods and achieve promising results. We briefly review some knowledge graph embedding methods and discuss their connections to our work.

\vspace{0.1in}
\textbf{Translation Distance Based Methods.} TransE~\citep{TransE} represents entities and relations as vectors and models the relation as a translation from the head entity to the tail entity, i.e., $\text{vec}(head) + \text{vec}(relation) = \text{vec}(tail)$. Along the line, TorusE~\citep{TorusE} and RotatE~\citep{sun2018rotate} are proposed to model relation as translation on a compact Lie group and rotation in a complex vector space, respectively. The relational operations in previous translation distance based methods are injective. The non-injective property of relations is discussed in several extensions of TransE, such as TransH~\citep{TransH} and TransR~\citep{TransR}. They project the vectors of entities into a subspace and then perform relational translation between entities in the subspace. Different entity vectors of non-injective relations could be the same in the subspace. Despite the capability of these methods to model non-injective relations, their overall performance on benchmark datasets lags behind the recent state-of-the-art methods such as RotatE~\citep{sun2018rotate} and ConvE~\citep{convE}, and they can only model part of the relation patterns.
MQuadE~\citep{MQuadE} addresses the problem of learning the non-injective relationships using a quadruple matrix representation for fact triples, in which the entity embedding matrices are required to be symmetric.
The MQuadE has good theoretical properties and performs well in real-world tasks.
Our proposed method - SpaceE has similar theoretical properties as MQuadE, and it does not impose symmetry on
the entity embedding matrices (in fact, the entity embedding matrices are not necessarily square),
which makes SpaceE more flexible and achieves comparable or even better performance than MQuadE in real-world tasks.

%Our model allows non-injective transformations with singular matrices for better non-injective relation modeling.

\vspace{0.1in}

\textbf{Bilinear Semantic Matching Methods.} RESCALE~\citep{Rescale} is the first bilinear model that uses matrices to represent relations. DistMult~\citep{DistMult2015} simplifies RESCALE and employs a diagonal matrix for relation modeling. As DistMult uses a symmetric score function, it cannot model skew-symmetry relations. \citet{trouillon2016complex} proposed ComplEx to model skew-symmetry relations.

\vspace{0.1in}
\textbf{Deep Learning Methods.} Neural networks such as neural tensor networks~\citep{tensorNN} and convolutional neural networks~\citep{convE} were leveraged for knowledge graph completion. Although they have strong expressiveness, they lack the interpretability of relation reasoning. Moreover, the high performances of some recent neural network methods~\citep{nathani-etal-2019-learning} can be attributed to the inappropriate evaluation protocols~\citep{sun-etal-2020-evaluation}.

\vspace{0.1in}
\textbf{Tensor Decomposition Methods.} The knowledge graph triples can be regarded as a 3-order tensor. Canonical Polyadic (CP) tensor decomposition is leveraged in~\citet{CPTensor, simpleIE} for knowledge graph completion. \citet{tuckerTensor} proposed TuckER based on Tucker tensor decomposition. TuckER benefits from multi-task learning by sharing parameters through the core tensor.

\section{Relation Modeling by Linear Transformation}
\vspace{0.1in}

\subsection{Notations} 
We use bold upper case letters to denote matrices. We denote the identity matrix as $\bm{I}$, the  Frobenius norm of a matrix $\bm{A}$ as $\|\bm{A}\|_{F}$, the element-wise product of two matrices $\bm{A}$ and $\bm{B}$ as $\bm{A} \odot \bm{B}$, and the Hadamard product of two complex vectors $\bm{a}$ and $\bm{b}$ as $\bm{a}\circ \bm{b}$.

\subsection{The SpaceE Model}
Let $\mathcal{E} = \{e_1, e_2, \cdots, e_n\}$ be the set of entities, $\mathcal{R} = \{r_1, r_2, \cdots, r_m\}$ be the set of relations, and $\mathcal{T} = \{(h_i, r_i, t_i)\}$  be the collection of fact triples, where $h_i \in \mathcal{E}$ is the head entity, and $t_i \in \mathcal{E}$ is the tail entity, $r_i \in \mathcal{R}$ is the relation.

%\newpage

SpaceE represents entities as $p \times q$ matrices and relations as $q \times q$ square matrices.  Let $\bm{H}, \bm{R}, \bm{T}$ denote the matrix of the head entity $h$, the relation $r$, and the tail entity $t$, respectively. SpaceE uses the following score function to measure the plausibility of a fact triple $(h, r, t)$: 
\begin{equation}\label{eq:SpaceE_score}
    d(h, r, t) = \| \bm{H} \bm{R}  - \bm{T} \|_{F}^2.
\end{equation}
The idea behind the score function is that the relation between two entities corresponds to a linear transformation of entity matrices. It is expected that $\bm{H}\bm{R}  \approx \bm{T}$ when the fact triple $(h, r, t)$ holds, while $\bm{H}\bm{R}$ should be far away from $\bm{T}$ otherwise. 

Since $(h, r, t)$ is equivalent to $(t, \hat{r}, h)$ where $\hat{r}$ is the reverse relation of relation $r$, we can adopt the reciprocal learning approach~\citep{CPTensor} and develop following score function:
\begin{equation} \label{eq:SpaceE_tail_score}
    f(h, r, t) = \|\bm{T} \bm{\hat{R}} - \bm{H}\|_{F}^2, 
\end{equation}
where $\bm{\hat{R}}$ is the representation of $\hat{r}$. The score function $d(h, r, t)$ is utilized for head entity prediction while the score function $f(h, r, t)$ is utilized for tail entity prediction during training and inference.
 As a result, a tail entity may have many head entities when $\bm R$ is singular, and a head entity may have many tail entities when $\bm{\hat{R}}$ is singular. In other words, the non-injective property of relation is preserved. See Section~\ref{sec:model} for more details.

\subsection{Training}
The training process samples a mini-batch of fact triples from training data and randomly corrupts the head entities and tail entities to obtain negative samples for head entity prediction and tail entity prediction,  respectively. It optimizes a loss function to make the scores of true triples lower than that of negative triples. After training, the scores are utilized to rank candidate entities to answer the link prediction queries.

We use the self-adversarial negative sampling loss function~\citep{word2vec, sun2018rotate} to learn the parameters:
\begin{equation*}
\begin{split}
\mathcal{L} =& -\log \sigma(\gamma - d(h, r, t)) -\log \sigma(\gamma - f(h, r, t)) \\
& - \sum_{i=1}^k p(\hat{h}_i, r, t) \log  \sigma(d(\hat{h}_i, r, t) - \gamma) 
 - \sum_{i=1}^k p(h, r, \hat{t}_i) \log \sigma( f(h, r, \hat{t}_i) - \gamma),
\end{split}
\end{equation*}
where $\sigma$ is the sigmoid function, $\gamma$ is the fixed margin, $\hat{h}_i$ is the $i$-th sampled negative head entity, and $\hat{t}_i$ is the $i$-th sampled negative tail entity, $p(\hat{h}_i, r, t)$ and $p(h, r, \hat{t}_i)$ are the negative sample weights in the self-adversarial loss~\citep{sun2018rotate} 
%, $p(\hat{h}_i, r, t_i)$ is the current prediction score 
defined as:
\begin{equation*}
p(\hat{h}_i, r, t) = \frac{ \exp \alpha ( \gamma - d(\hat{h}_i, r, t) )}{\sum_{j=1}^k \exp \alpha (\gamma - d(\hat{h}_j, r, t))},
\qquad
p({h}, r, \hat{t}_i) = \frac{ \exp \alpha ( \gamma - f(h, r, \hat{t}_i) )}{\sum_{j=1}^k \exp \alpha (\gamma - f(h, r, \hat{t}_j))},
\end{equation*}
where $\alpha$ is the self-adversarial temperature. The weight corresponding to a negative sample is large when the model predicts the negative sample as a true fact triple.

The orthogonal constraint can stabilize the training and ease overfitting~\citep{bansal2018orthogonality}. Orthogonal linear transformation has the nice property of preserving the norm. We propose a regularization term to encourage the relation matrix to be nearly orthogonal. Note that orthogonal matrices $\bm{R}^T \bm{R} = \bm{I}$ and they are non-singular. We design a regularization term $\| (\bm{R}^T \bm{R}) \odot (\bm{R}^T \bm{R}) - \bm{R}^T \bm{R} \|_{F}^2$ to encourage the elements in $\bm{R}^T \bm{R}$ to be either 1 or 0. The relation matrix satisfying the regularization constraint is nearly orthogonal and could be singular. The regularized loss function is 
\begin{equation*}
\begin{split}
    \mathcal{L}_{reg} =  \mathcal{L} + \lambda (\| (\bm{R}^T \bm{R}) \odot (\bm{R}^T \bm{R}) - \bm{R}^T \bm{R} \|_{F}^2 
    + \| (\bm{\hat{R}}^T \bm{\hat{R}}) \odot (\bm{\hat{R}}^T \bm{\hat{R}}) - \bm{\hat{R}}^T \bm{\hat{R}} \|_{F}^2),
\end{split}
\end{equation*}
where $\lambda$ is the regularization hyper-parameter.

\section{Model Property Analysis}\label{sec:model}

\vspace{0.1in}

\subsection{Infer Relation Properties}

\begin{theorem} \label{theorem::spaceE}
SpaceE can model non-injective, symmetric, skew-symmetric, inversion, Abelian composition, and non-Abelian composition relations with different relation matrices, as summarized in Table~\ref{tab::SpaceE_property}. 
\end{theorem}

\begin{table}[ht]
\centering
\begin{tabular}{ll}
\hline
\textbf{Relation Property}                & \textbf{Relation Matrix}  \\ \hline
Non-injective                             & $\bm{R}$ or $\hat{\bm{R}}$ is singular \\
Symmetric                        & $\bm{R}^2 = \bm{I}$                 \\
Skew-symmetric                   & $\bm{R}^2 \neq \bm{I}$              \\
$r_2$ inversion of $r_1$                  & $\bm{R}_1 \bm{R}_2 = \bm{R}_2 \bm{R}_1 = \bm{I}$   \\
 $r_3 = r_1 \otimes r_2 $      & $\bm{R}_3 = \bm{R}_1 \bm{R}_2 $           \\
$r_1 \otimes r_2 $ Abelian     & $ \bm{R}_1 \bm{R}_2 = \bm{R}_2 \bm{R}_1$       \\
$r_1 \otimes r_2$ Non-Abelian  & $\bm{R}_1 \bm{R}_2 \neq \bm{R}_2 \bm{R}_1$  \\
\hline
\end{tabular}
\caption{The ability of SpaceE to model various relation properties with different relation matrices.}
\label{tab::SpaceE_property}
\end{table}

\textit{Proof.} 
\textbf{Non-injective Relation.} A relation $r$ is non-injective, iff there exist multiple fact triples $(h_1, r, t), \cdots, (h_k, r, t)$, where $h_1, \cdots, h_k$ are different or multiple fact triples $(h, r, t_1), \cdots, (h, r, t_k)$, where $t_1, \cdots, t_k$ are different. In SpaceE, when $\bm{R}$ is singular, the equation $\bm{X}\bm{R} = \bm{T}$ can have different solutions $\bm{X} = \bm{H_1}, \cdots,$ $\bm{H_k}$; when $\bm{\hat{R}}$ is singular, the equation $\bm{Y}\bm{\hat{R}} = \bm{H}$ can have different solutions $\bm{Y} = \bm{T_1}, \cdots, \bm{T_k}$.

\vspace{0.1in}

\textbf{Symmetric Relation.} For a symmetric relation $r$, for any two entities $h, t \in \mathcal{E}$,  the fact triple $(h ,r ,t)$ holds $\iff$ the fact triple $(t, r, h)$ holds.
In SpaceE, it requires
\begin{equation*}
 \bm{H}\bm{R} = \bm{T} \quad \iff \quad  \bm{T}\bm{R} = \bm{H}.
\end{equation*}
It follows from $\bm{R}^2 = \bm{I}$.

\vspace{0.1in}

\textbf{Skew-symmetric Relation.} For a skew-symmetric relation r,  for any two entities $h, t \in \mathcal{E}$, the fact triple $(h, r, t)$ is true $\implies$ the fact triple $(t, r, h)$ is false.  This requires in SpaceE that
\begin{equation*}
     \bm{H}\bm{R} = \bm{T} \quad \implies \quad  \bm{T}\bm{R} \neq \bm{H}.
\end{equation*}
It follows from $\bm{R}^2 \neq \bm{I}$.

\vspace{0.1in}

\textbf{Inversion Relation.} A relation $r_2$ is the inversion of relation $r_1$ iff for any two entities $h, t \in \mathcal{E}$, the fact triple $(h, r_1, t)$ is true $\iff$ the fact triple $(t, r_2, h)$ is true.  It requires in SpaceE, 
\begin{equation*}
     \bm{H}\bm{R_1} = \bm{T} \iff  \bm{T}\bm{R_2} = \bm{H}.
\end{equation*}
It follows from $\bm{R_1} \bm{R_2} = \bm{R_2} \bm{R_1} = \bm{I}$.

\vspace{0.1in}

\textbf{Relation Composition.} A relation $r_3 = r_1 \otimes r_2 $ is the composition of two relations $r_1$ and $r_2$ iff for any three entities $a, b, c  \in \mathcal{E}$,  the facts $(a, r_1, b), (b, r_2, c)$ are true $\implies$ the fact $(a, r_3, c)$ is true. This requires in SpaceE that
\begin{equation*}
      \bm{A}\bm{R_1} = \bm{B},  \bm{B}\bm{R_2} = \bm{C} \implies  \bm{A}\bm{R_3}  = \bm{C}
\end{equation*}
It follows from $\bm{R_3} = \bm{R_1}\bm{R_2}$.

\vspace{0.1in}

\textbf{Abelian Composition and Non-Abelian Composition.} Let $r_3 = r_1 \otimes r_2$ be the composition of the relation $r_1$ and $r_2$, $r_4 = r_2 \otimes r_1$ be the composition of the relation $r_2$ and $r_1$.  In SpaceE, the matrix representations of the two composite relations $r_3$ and $r_4$ can be written as 
\begin{equation*}
\bm{R_3} =  \bm{R_1}\bm{R_2},  \quad \bm{R_4} = \bm{R_2}\bm{R_1}.    
\end{equation*}
When $\bm{R_1} \bm{R_2} = \bm{R_2}\bm{R_1}$, the composition of $r_1$ and $r_2$ is Abelian; otherwise, the composition is non-Abelian.

\subsection{Connection to RotatE}\label{section::linkRotatE}
We show that our model SpaceE can be regarded as an extension of RotatE~\citep{sun2018rotate}.

\begin{theorem} \label{theorem::connection}
SpaceE subsumes RotatE with the special block diagonal matrix representations of complex vectors.
\end{theorem}

\textit{Proof.} \  RotatE uses complex vectors to represent entities and relations and models the relation between two entities as the rotation in complex vector space. The score function of a fact triple $(h, r, t)$ in RotatE is
\begin{equation} \label{eq::rotate_score}
    s(h, r, t) = \|\bm{h} \circ \bm{r} - \bm{t} \|, \quad \quad |\bm{r}_i| = 1, 
\end{equation}
where $\bm{h}, \bm{r}, \bm{t} \in \mathbb{C}^K$ are complex vector representations of head entity, relation, tail entity respectively.

\vspace{0.1in}

A complex number $z = a + bi$ can be represented by a $2 \times 2$ matrix $
\begin{pmatrix}
a & -b \\
b & a 
\end{pmatrix}
$.  The product of two complex numbers $z_1 = a + bi$ and $z_2 = c + di$ is $z_1z_2 = ac -bd + (bc + ad)i$, which corresponds to the product of two matrices:
\begin{equation*}
\begin{pmatrix}
a & -b \\
b & a 
\end{pmatrix}
\begin{pmatrix}
c & -d \\
d & c 
\end{pmatrix} = 
\begin{pmatrix}
ac - bd & -ad -bc \\
bc + ad & -bd + ac
\end{pmatrix}.
\end{equation*}

\vspace{0.1in}

A complex vector $\bm{z} = (z_1, z_2, \cdots, z_K) \in \mathbb{C}^K$ can be represented as a $2K \times 2K $ block diagonal matrix $\bm{Z} = \text{diag} \{\bm{Z}_1, \bm{Z}_2, \cdots, \bm{Z}_K \}$ whose $i$-th block component $\bm{Z}_i$ is the $2 \times 2$ matrix representation of $z_i$. Let $\bm{z}, \bm{s} \in \mathbb{C}^K$ be two complex vectors and $\bm{Z}$, $\bm{S}$ be their corresponding block diagonal matrices respectively. The Hadamard product of  $\bm{z}$ and $\bm{s}$ is
\begin{equation*}
    \bm{z} \circ \bm{s} = (z_1s_1, z_2s_2, \cdots, z_Ks_K).
\end{equation*}
It is equivalent to the product of matrices $\bm{Z}$ and $\bm{S}$,
\begin{equation*}
    \bm{Z}\bm{S} = \text{diag}\{\bm{Z}_1\bm{S}_1, \bm{Z}_2\bm{S}_2, \cdots, \bm{Z}_K\bm{S}_K\}.
\end{equation*}
As a result, the score function of RotatE corresponds to the score function of SpaceE if we change the complex vector representations of entities and relations to the corresponding block diagonal matrix representations.

\vspace{0.2in}
\subsection{Time and Space Complexity} \label{section::parameter_num}
\vspace{0.2in}

RotatE and TransE use vectors of dimension $n$, the time and space complexity is $\mathcal{O}(n)$. SpaceE uses $p \times q$ and $q \times q$ matrices for entity and relation representations, respectively. Its space complexity is $\mathcal{O}(pq + q^2)$; its time complexity is $\mathcal{O}(pq^2)$. Assume that there are $n_e$ entities and $n_r$ relations, the total number of parameters in SpaceE is $n_e pq + n_r q^2$. In experiments, we set $p \approx \sqrt{n}, q \approx \sqrt{n}$ and use the number of parameters not more than RotatE for a fair comparison. For example, in the FB15K dataset, RotatE uses 1000 dimensions complex vectors; it has 20000 parameters of a complex vector; 1000 for the real part, 1000 for the imaginary part. SpaceE uses $45\times45=2025$ dimensions embedding matrices. In the Yago3-10 dataset, RotatE has $500\times2=1000$ parameters in a complex vector; SpaceE has $20\times40=800$ parameters in an embedding matrix.

\newpage

\section{Experiments}

\vspace{0.1in}

\subsection{Experimental Setup}

\textbf{Datasets.} We conduct experiments on five benchmark datasets: FB15k, WN18, FB15K-237, WN18RR, and YAGO3-10. The statistics of the datasets are summarized in Table~\ref{tab::dataset}.

\vspace{0.2in}

\begin{table*}[h]
\centering
\begin{tabular}{llllllll}
\hline
\textbf{Dataset} & \textbf{\#entity} & \textbf{\#relation} & \textbf{\#train} & \textbf{\#valid} & \textbf{\#test}  & \textbf{\#\textit{tphr}} & \textbf{\#\textit{hptr}} \\ \hline
FB15k            & 14951      & 1345       & 483142           & 50000            & 59071 & 9.3 & 19.7           \\ 
WN18             & 40943      & 18         & 141442           & 5000             & 5000 & 4.2 & 4.2            \\ 
FB15K-237        & 14541      & 237        & 272115           & 17535            & 20466 & 7.9 & 65.9 \\ 
WN18RR           & 40943      & 11         & 86835            & 3034             & 3134 & 4.5 & 2.9 \\ 
YAGO3-10         & 123182     & 37         & 1079040          & 5000             & 5000 & 3.5 & 913.1 \\ \hline
\end{tabular}
\caption{\label{tab::dataset}
Statistics about the datasets. Among all the triplets of a
relation $r$, let \textit{tphr} denote the
average number of tail entities per head entity, \textit{hptr} denote the average number of head entities per tail entity. \#\textit{tphr} denotes the average of \textit{tphr} for all relations, \#\textit{hptr} denotes the average of \textit{hptr} for all relations.}\vspace{0.2in}
\end{table*}

FB15k~\citep{TransE} is a subset of FreeBase~\citep{freebase2008}, a large-scale knowledge graph about the general world knowledge. FB15k samples 15k entities and their relations such as \textit{/location/country/capital} from FreeBase.

WN18~\citep{TransE} is a subset of the WordNet,  a lexical database for the English language that groups synonymous words into synsets. WN18 contain relations between words such as \textit{hypernym} and \textit{similar\_to}.

FB15k-237~\citep{FB15k-237} and WN18RR~\citep{convE} are subsets of FB15k and WN18, respectively, with the inverse relation deleted to resolve the test set leakage problem and to examine the relation composition modeling ability.

YAGO3-10~\citep{Yago3} is a subset of YAGO3 whose entities have at least ten relations. Most of the relations in YAGO3-10 are descriptive attributes of people such as \textit{wasBornIn}, \textit{worksAt}, and \textit{graduatedFrom}, which are non-injective. As shown in Table~\ref{tab::dataset}, the average number of head per tail of these relations is 913.

\vspace{0.15in}

\textbf{Evaluation Metrics and Protocols.} We use the mean reciprocal rank (MRR) and HIT@1, 3, 10 as the metrics to evaluate different models. MRR measures the average of the inverse rank of correct entities in the list predicted by the model. HIT@k measures the average percentage of correct entities that are ranked in the top k by the model. Following~\citet{TransE}, we use the \textit{filtered} evaluation setting where the triplets that appear either in the training, validation, or test set (except the test triplet of interest) are removed from the list of corrupted triplets. To deal with the case that the model may predict the same score for different fact triples, we adopt the \textbf{random evaluation protocol} suggested by~\citet{sun-etal-2020-evaluation}.

\vspace{0.15in}

\textbf{Baselines.} We compare our model with representative state of the art models, including distance translation based methods (TransE~\citep{TransE}, SE~\citep{bordes2011SE}, TransH~\citep{TransH}, RotatE~\citep{sun2018rotate}, BoxE~\citep{BoxE}), bi-linear semantic matching methods (DistMult~\citep{DistMult2015}, ComplEx~\citep{trouillon2016complex}, 
TuckER~\citep{tuckerTensor}), and a deep learning method ConvE~\citep{convE}.

\vspace{0.15in}

\textbf{Implementation Details.} The Adam optimizer~\citep{adam} is utilized for model training. We tune the hyper-parameters with grid search and select the model that have highest MRR on the validation dataset. The hyper-parameters are selected from following configurations: the dimension of entity and relation matrices $p, q\in \{10, 20, 40,  45\}$,  the self-adversarial temperature $\alpha \in \{0.5, 1.0\}$, the fixed margin $\gamma \in \{3, 6, 9, 12, 24\}$, the number of negative samples $k \in \{256, 512, 1024\}$, the regularization coefficient $\lambda \in \{0.005, 0.01, 0.05, 0.1, 0.3, 0.6\}$, the batch size $b \in \{512, 1024\}$, the initial learning rate $lr \in \{1e-4, 2e-4\}$. The best hyper-parameters on each dataset are given in Table~\ref{tab::hyperparameters}. The entity and relation matrix parameters are randomly initialized from the normal distribution $\mathbb{N}(0, 0.01)$. Our algorithms are implemented on the PaddlePaddle deep learning framework \url{https://www.paddlepaddle.org.cn/}.

\begin{table}[t]
\centering
\begin{tabular}{c|ccccc}
\hline
          & FB15k & WN18 & FB15k-237 & WN18RR & YAGO3-10 \\ \hline
$p$       & 45    & 45   & 45        & 10     & 20       \\
$q$       & 45    & 45   & 45        & 40     & 40       \\
$b$       & 1024  & 512  & 1024      & 512    & 1024     \\
$\alpha$  & 1     & 0.5  & 1         & 0.5    & 1        \\
$\gamma$  & 24    & 12   & 9         & 3      & 12       \\
$k$       & 256   & 1024 & 256       & 1024   & 256      \\
$lr$      & 1e-4  & 1e-4 & 1e-4      & 2e-4   & 2e-4     \\
$\lambda$ & 0.6   & 0.05 & 0.05      & 0.1    & 0.005   \\ \hline
\end{tabular}
\caption{The best hyper-parameters on benchmark datasets. Hyper-parameters $p, q, b, \alpha, \gamma, k, lr, \lambda$, respectively, represent the number of rows of the embedding matrix, the number of columns of the embedding matrix, the batch size, the self-adversarial temperature in negative sampling, the fix margin in the loss function, the number of negative samples, the initial learning rate of the optimizer, the regularization coefficient.} \label{tab::hyperparameters}
\end{table}

\subsection{Main Results}

We highlight the best results on the model category with a bold font and mark the best overall results by a surrounding rectangle in Tables~\ref{tab::fb15k},~\ref{tab::fb15k-237},~\ref{tab::yago}.

Table~\ref{tab::fb15k} summarizes the results on the FB15k and WN18 datasets. The results of TransE are taken from later work's re-implementation~\citep{nickel2016holographic}. The results show that SpaceE can get comparative results on all the evaluation metrics with the state of the art baselines. TransH explicitly models non-injective relations but does not get overall competitive results. As discussed in~\citet{convE, sun2018rotate},  many test triples in the two datasets appear as a reciprocal form of the training samples, and the test set leakage through inverse relation makes the datasets less challenging. The main relation patterns of the two datasets are symmetry, skew-symmetry, and inversion. The competitive performance of SpaceE demonstrates its capability to model the three relation patterns.

\vspace{0.1in}

\begin{table*}[ht]
	\centering
		\begin{tabular}{l|llll|llll}
			\hline
			\multirow{2}{*}  & \multicolumn{4}{c|}{\textbf{FB15k}} & \multicolumn{4}{c}{\textbf{WN18}} \\ \cline{2-9}
			& MRR     & H@1    & H@3    & H@10    & MRR     & H@1    & H@3    & H@10   \\ \hline
			
			DistMult[\markcir]  & \circletext{\textbf{0.798}}   & -      & -      & \circletext{\textbf{0.893}}   & 0.797   & -      & -      & 0.946  \\ 
			ComplEx & 0.692 & 0.599 & 0.759 & 0.840 & \textbf{0.941} & \textbf{0.936} & \textbf{0.945} &  \textbf{0.947} \\  \hline 
	   % DihEdral D4-STE  & 0.733 & 0.641 & 0.803 & 0.877 & 0.946 & 0.942 & 0.948 & 0.952 \\ \hline 		
        
        ConvE     & 0.657   & 0.558  & 0.723  & 0.831   & 0.943   & 0.935  & 0.946  & 0.956  \\ \hline
			SE &  - & - & - & 0.398 & - & - & - & 0.805 \\ 
			TransE[\marksqure]   & 0.463   & 0.297  & 0.578  & 0.749   & 0.495   & 0.113  & 0.888  & 0.943  \\ 
			TransH    & -         &     -   & -        & 0.585        &    -     &    -    &   -     &  0.867      \\

			RotatE    & 0.797   & \circletext{\textbf{0.746}}  & \circletext{\textbf{0.830}}  & 0.884  & \circletext{\textbf{0.949}}  & \circletext{\textbf{0.944}}  & \circletext{\textbf{0.952}}  & \circletext{\textbf{0.959}} \\ 
			SpaceE  & 0.791        &   0.736     &   \circletext{\textbf{0.830}}     &     0.883    & 0.947        &     0.941   &  0.951     &   \circletext{\textbf{0.959}}    \\ 
			\hline
		\end{tabular}
			\caption{\label{tab::fb15k}
		Results on FB15k and WN18 datasets. Results with suffix [\marksqure] and  [\markcir] are taken from~\citet{nickel2016holographic} and~\citep{baselineBack} respectively. Others are obtained from the original papers.}
\end{table*}

The difference between the results of RotatE and SpaceE is minor. This could be explained by our analysis of the connection between SpaceE and RotatE in Section~\ref{section::linkRotatE}.

%\newpage

\begin{table*}[ht]
	\centering
		\begin{tabular}{l|llll|llll}
			\hline
			\multirow{2}{*}{} & \multicolumn{4}{c|}{\textbf{FB15k-237}} & \multicolumn{4}{c}{\textbf{WN18RR}}   \\ \cline{2-9} 
			& MRR   & H@1   & H@3   & H@10  & MRR   & H@1  & H@3  & H@10   \\ \hline
			DistMult         & 0.24  & 0.155 & 0.263 &  0.419 & 0.43     &   0.39    &   0.44   &  0.49 \\ 
			
	ComplEx & 0.247 & 0.158 & 0.275 & 0.428 & 0.44 & 0.41 & 0.46 & 0.51 \\
			TuckER & \circletext{\textbf{0.358}} & \circletext{\textbf{0.266}} & \circletext{\textbf{0.394}} & \circletext{\textbf{0.544}} & \textbf{0.470} & \circletext{\textbf{0.443}} & \textbf{0.482} & \textbf{0.526} \\ \hline
	        
	        ConvE             & 0.325 & 0.237 & 0.356  & 0.501 & 0.43      &  0.40  & 0.44   &  0.52 \\ \hline
			TransE            & 0.294 & -      & -       &  0.465     &   0.226    &      - &   -   &     0.501    \\ 
			RotatE            & 0.338      &    0.24   &  0.375      & 0.533      &    \circletext{\textbf{0.476}}   & \textbf{0.428}     & 0.492     & \circletext{\textbf{0.571}} \\ 
			BoxE & 0.337 & - & - & 0.538 & 0.451 & - & - & 0.541 \\ 
			SpaceE           &  \textbf{0.351}     & \textbf{0.253}      &  \textbf{0.389}     &  \circletext{\textbf{0.544}}      &   0.473    &   0.423   &   \circletext{\textbf{0.496}}   &   0.570 \\ 
			\hline
		\end{tabular}
	\caption{\label{tab::fb15k-237}
		Results on FB15k-237 and WN18RR datasets. The results of TransE, DistMult, and ConvE are taken from~\citet{sun2018rotate}, and others are obtained from the original papers. SE~\citep{bordes2011SE} and TransH~\citep{TransH} papers do not have results on the two datasets.
	}
\end{table*}

From the results in Table~\ref{tab::fb15k-237}, we observe that on FB15k-237, SpaceE achieves the best performance among the translation distance based knowledge graph embedding methods on all evaluation metrics. Its performance is competitive with the state-of-the-art semantic matching method TuckER, especially on Hits@10. The main relation pattern in Fb15k-237 is composition. The results demonstrate the effectiveness of SpaceE to model the relation composition.

\begin{table}[htbp]
    \centering
    \begin{tabular}{l|llll}
    \hline
    \multirow{2}{*}{} & \multicolumn{4}{c}{\textbf{YAGO3-10}} \\ \cline{2-5}
    & MRR   & H@1   & H@3   & H@10 \\ \hline
    DistMult & 0.340 & 0.240  & 0.380 & 0.540 \\
    ComplEx & 0.360 & 0.260 & 0.40 & 0.550 \\ 
    TuckER &  \textbf{0.527} & \textbf{0.446} & \textbf{0.576} & \textbf{0.676} \\ \hline
    ConvE & 0.52 & 0.45 &  0.56   & 0.66 \\ \hline
    RotatE & 0.495 & 0.402 & 0.550 & 0.670 \\
    BoxE & \circletext{\textbf{0.560}} & - & - & 0.691 \\
    MQuadE & 0.536  & 0.449 & 0.592 & 0.689 \\
    SpaceE & 0.549 & \circletext{\textbf{0.463}} & \circletext{\textbf{0.604}} & \circletext{\textbf{0.702}} \\ \hline
    \end{tabular}
    \caption{Results on YAGO3-10 datasets. The results of DistMult is taken from~\citet{sun2018rotate}. Others are obtained from the origin papers.}
    \label{tab::yago}
\end{table}

The results in Table~\ref{tab::yago} show that SpaceE gets the state-of-the-art performance on Hits@1, Hits@3, and Hits@10 metrics on YAGO3-10. The results are promising since YAGO3-10 is the largest dataset among the three datasets. It contains 123182 entities and more than one million fact triples. Most of the relations in YAGO3-10 are non-injective descriptive attributes of people such as \textit{wasBornIn} and \textit{graduatedFrom}. RotatE does not perform well on this dataset due to its limited ability for non-injective relation modeling. The high performances of SpaceE demonstrate its strong capability of non-injective relation modeling.

As shown in Table~\ref{tab::fb15k-237}, SpaceE gets competitive results with RotatE on WN18RR. RotatE performs well on this dataset, although not well on FB15k-237 and YAGO3-10 datasets. We find two major reasons for this phenomenon. First, the non-injective relations which are difficult for RotatE are not prevalent in WN18RR. \#\textit{hptr} of relations in WN18RR is 2.9. Second, symmetry is prevalent in WN18RR. RotatE can take advantage of its bias - the composition of two symmetric relations is (incorrectly) symmetric~\citep{BoxE}.

One might wonder whether adding the augmented inverse relations and their embeddings can improve RotatE. We conduct the experiments and get almost the same performances as RotatE; for example, the MRR metrics on FB5k-237 and YAGO3-10 are 0.338 and 0.506, respectively.

\subsection{Results on Non-injective Relations}

We compare our model with RotatE, the state-of-the-art translation distance based method on non-injective relations. Following previous works~\citep{TransH, sun2018rotate}, we study the ability to model the non-injective relations by categorizing relations into 1-to-1, 1-to-N, N-to-1, and N-to-N relations and report the performances of the four relation groups. 

The FB15k-237 test set has 74 1-to-1 relation triples, 67 1-to-N relation triples, 1710 N-to-1 relation triples, and 18615 N-to-N relation triples. The Yago3-10 test set has 556 N-to-1 relation triples and 4444 N-to-N relation triples; it does not contain 1-to-1 and 1-to-N relation triples.

The results in Table~\ref{tab::fb15k-237::rel_type} and Table~\ref{tab::yago3::rel_type} show that SpaceE consistently outperforms RotatE on N-N relations for both head and tail entity prediction, on 1-to-N relations for tail entity prediction, and on N-to-1 relations for head entity prediction.

\begin{table*}[t]
\centering
\begin{tabular}{c|c|cc|cc|cc|cc}
\hline
\multirow{2}{*}{Metric} & \multirow{2}{*}{Model} & \multicolumn{2}{c|}{1-1} & \multicolumn{2}{c|}{1-N} & \multicolumn{2}{c|}{N-1} & \multicolumn{2}{c}{N-N} \\ \cline{3-10}
                        &                        & H          & T          & H          & T          & H          & T          & H          & T          \\ \hline
\multirow{2}{*}{MRR}    & RotatE                 & 1          & 1          & \textbf{0.885}      & 0.168      & 0.095      & \textbf{0.847}      & 0.243      & 0.389      \\ 
                        & SpaceE                 & 1          & 1          & 0.657      & \textbf{0.267}      & \textbf{0.152}      & 0.831      & \textbf{0.259} & \textbf{0.41}       \\ \hline
\multirow{2}{*}{H@10}   & RotatE                 & 1          & 1          & \textbf{0.985}      & 0.239      & 0.159      & \textbf{0.911}      & 0.441      & 0.610      \\ 
                        & SpaceE                 & 1          & 1          & \textbf{0.985}      & \textbf{0.582}      & \textbf{0.228}      & 0.905      & \textbf{0.455}      & \textbf{0.624} \\ \hline     
\end{tabular}
\caption{Results of RotatE and SpaceE on different relation types of the FB15k-237 dataset. H denotes head prediction performances, T denotes tail prediction performances.}
\label{tab::fb15k-237::rel_type}
\end{table*}

\begin{table}[t]
\centering
\begin{tabular}{c|c|cc|cc}
\hline
\multirow{2}{*}{Metric} & \multirow{2}{*}{Model} & \multicolumn{2}{c|}{N-to-1}     & \multicolumn{2}{c}{N-to-N}     \\ \cline{3-6} 
                        &                        & H           & T           & H           & T           \\ \hline
\multirow{2}{*}{MRR}    & RotatE                 & 0.01          & \textbf{0.66} & 0.38          & 0.64          \\ 
                        & SpaceE                & \textbf{0.02}  & 0.65          & \textbf{0.45} & \textbf{0.69} \\ \hline
\multirow{2}{*}{H@10}   & RotatE                 & 0.04          & 0.79           & 0.61          & 0.79          \\ 
                        & SpaceE                & \textbf{0.04} & \textbf{0.80}   & \textbf{0.65} & \textbf{0.81} \\ \hline
\end{tabular}
\caption{Head prediction and tail prediction results of RotatE and SpaceE on different relation types of the Yago3-10 dataset. H and T denote head prediction and tail prediction, respectively.}
\label{tab::yago3::rel_type}
\end{table}

The number of head entities attached to (t, r) (denoted by \textit{hptr}) or the number of tail entities attached to (h, r) (denoted by \textit{tphr}) can reflect the non-injective degree of the relation in a fact triple. The Yago3-10 dataset has the largest average number of heads per tail and relations (\textit{\#hptr}) among the five datasets, so we use it to investigate the relationship between the head prediction performance and \textit{hptr}. There are 18 triples in the test set which contain entities that do not appear in the training set and can not be predicted by any model. We filter them to avoid noises in the investigation.

We compare the head prediction performances of RotatE and SpaceE on test triples with different \textit{hptr}. The test triples are categorized into five groups by their \textit{hptr}:  [0, 10], [11, 50], [51, 100], [101, 1000], and [1001, 100000]. The average number of \textit{hptr} of the triples in the five groups is 4, 28, 74, 273, and 46678. The number of test triples in the five categories is 987, 913, 692, 2016, and 374.

\begin{table}[htbp]
	\centering
	\begin{tabular}{l|l|ccccc}
		\hline
		\multicolumn{2}{c|}{ \textit{hptr} group }&  1 & 2 & 3 & 4 & 5\\ \hline
		\multirow{2}{*}{MRR} & RotatE &  0.35 &  0.45 & 0.42 & 0.32 & 2e-3  \\
		 & SpaceE & \textbf{0.37} & \textbf{0.52} & \textbf{0.53} & \textbf{0.41} & \textbf{3e-3} \\ \hline
		 
		 \multirow{2}{*}{H@1} & RotatE & 0.27 & 0.34 & 0.28 & 0.18 & 0\\
	 & SpaceE & \textbf{0.30} & \textbf{0.42} & \textbf{0.41} & \textbf{0.28} & 0 \\ \hline
	 
	 	\multirow{2}{*}{H@3} & RotatE & 0.38 & 0.53 & 0.52 & 0.37 & 0 \\
	 	& SpaceE & \textbf{0.40} &  \textbf{0.59} & \textbf{0.60} & \textbf{0.47} & 0 \\ \hline
	 	
	 	\multirow{2}{*}{H@10} & RotatE & 0.48 & 0.64 & 0.69 & 0.60 & 3e-3 \\ 
	 	& SpaceE & \textbf{0.49} & \textbf{0.67} & \textbf{0.73} & \textbf{0.64} & \textbf{0.01} \\ 
	 	
		 \hline
	\end{tabular}
\caption{The head entity prediction performances of RotatE and SpaceE on triples with different \textit{hptr} in the YAGO3-10 dataset.} 
	\label{tab::test_category_prediction}\vspace{0.2in}
\end{table}

From the results in Table~\ref{tab::test_category_prediction}, we observe that: 1) SpaceE consistently outperforms RotatE in terms of all evaluation metrics on the five \textit{hptr} triple categories;  2) when \textit{hptr} $\leq$ 10, SpaceE beats RotatE with more than 0.01 absolute improvement; when 100 $\leq$ \textit{hptr} $\leq$ 1000, the improvement of SpaceE over RotatE is more substantial, with about 0.10 absolute improvement on MRR and Hits@1, 3.

\begin{figure}[htbp]

\mbox{\hspace{-0.3in}
    \begin{subfigure}{0.27 \textwidth}
     \includegraphics[width=\textwidth]{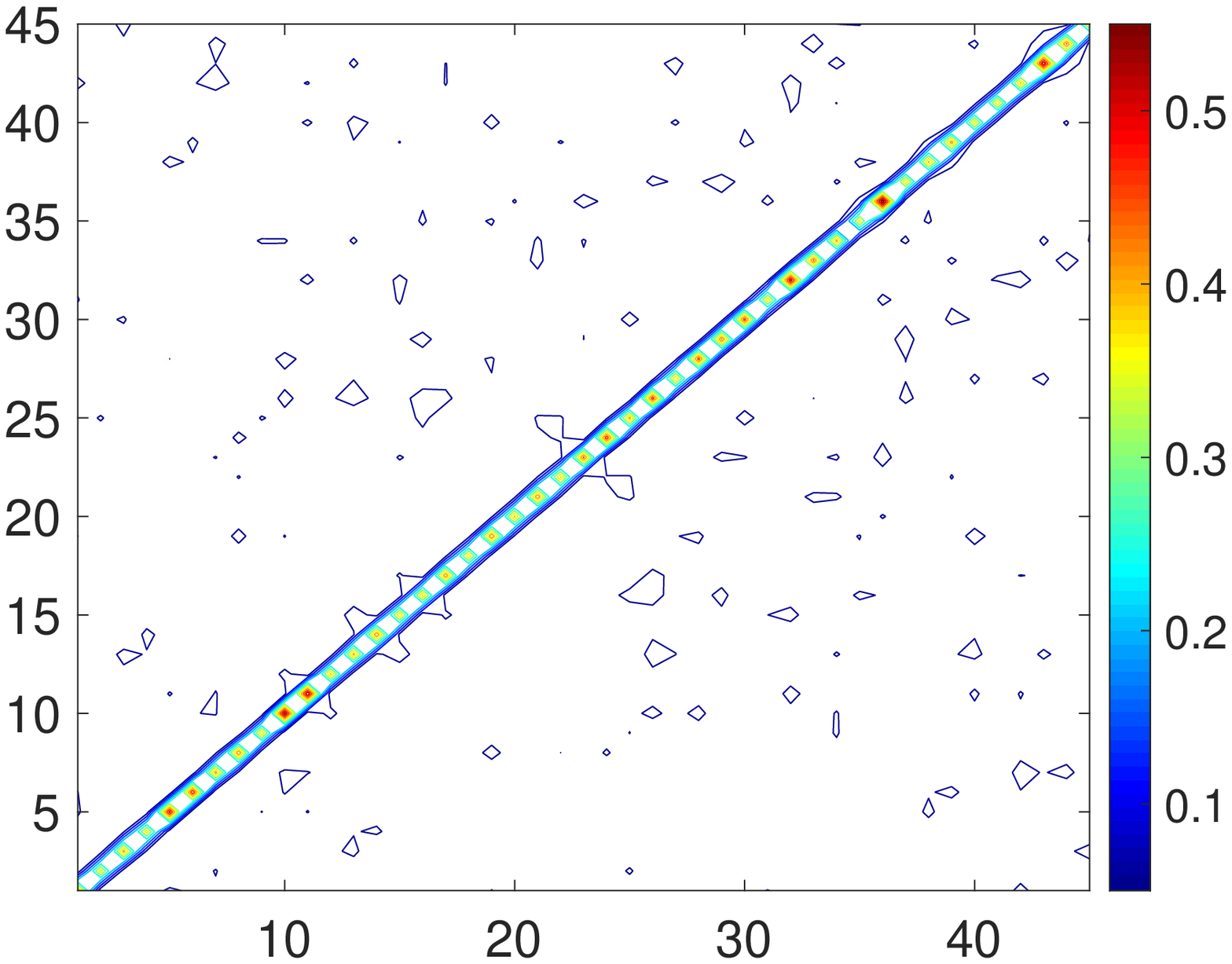}
     \caption{\textit{similar\_to}}
     \label{fig:similar}
    \end{subfigure} %
    \begin{subfigure}{0.27 \textwidth}
    \includegraphics[width=\textwidth]{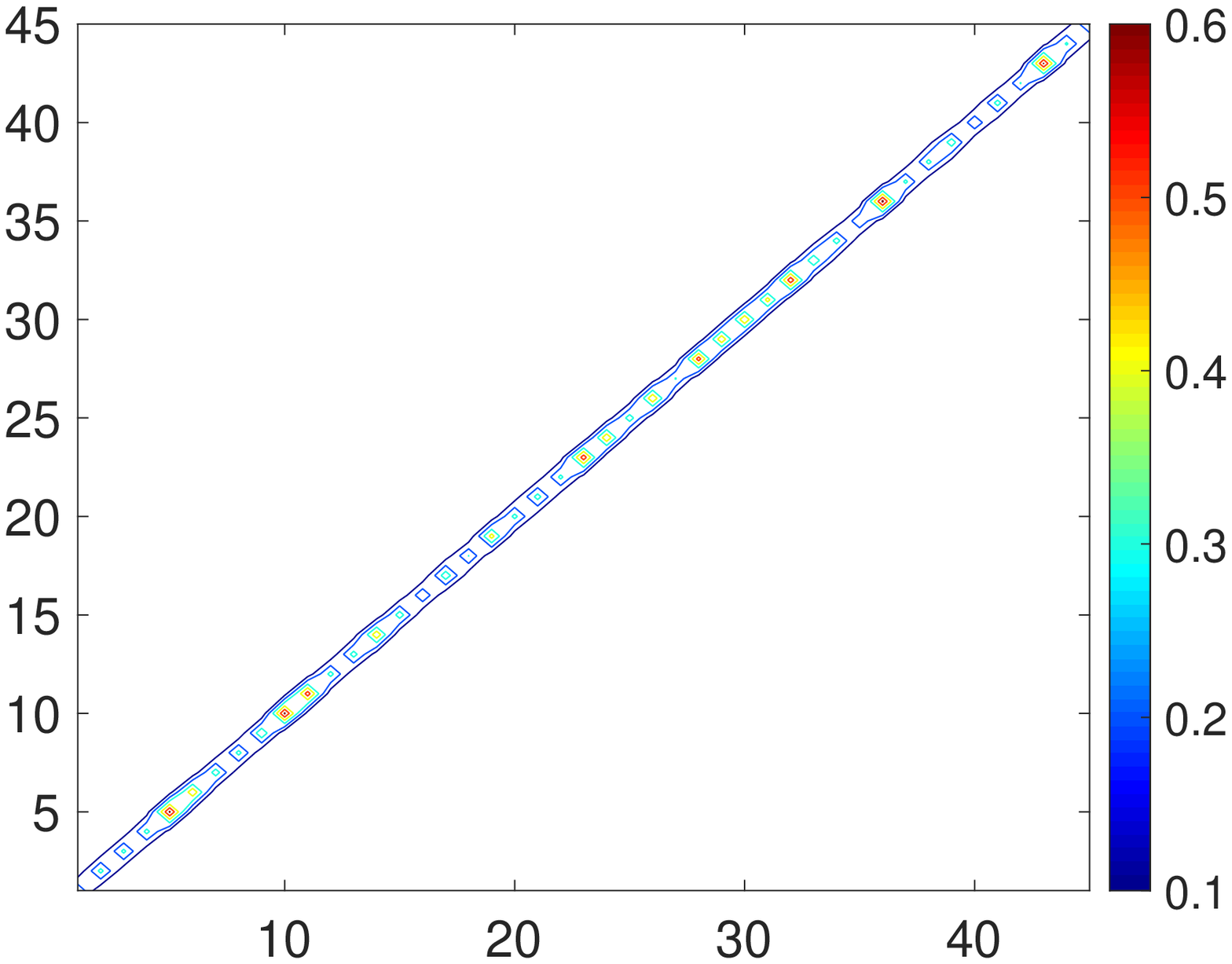}
    \caption{\textit{verb\_group}}
    \label{fig:verb}
    \end{subfigure} %
    \begin{subfigure}{0.27 \textwidth}
     \includegraphics[width=\textwidth]{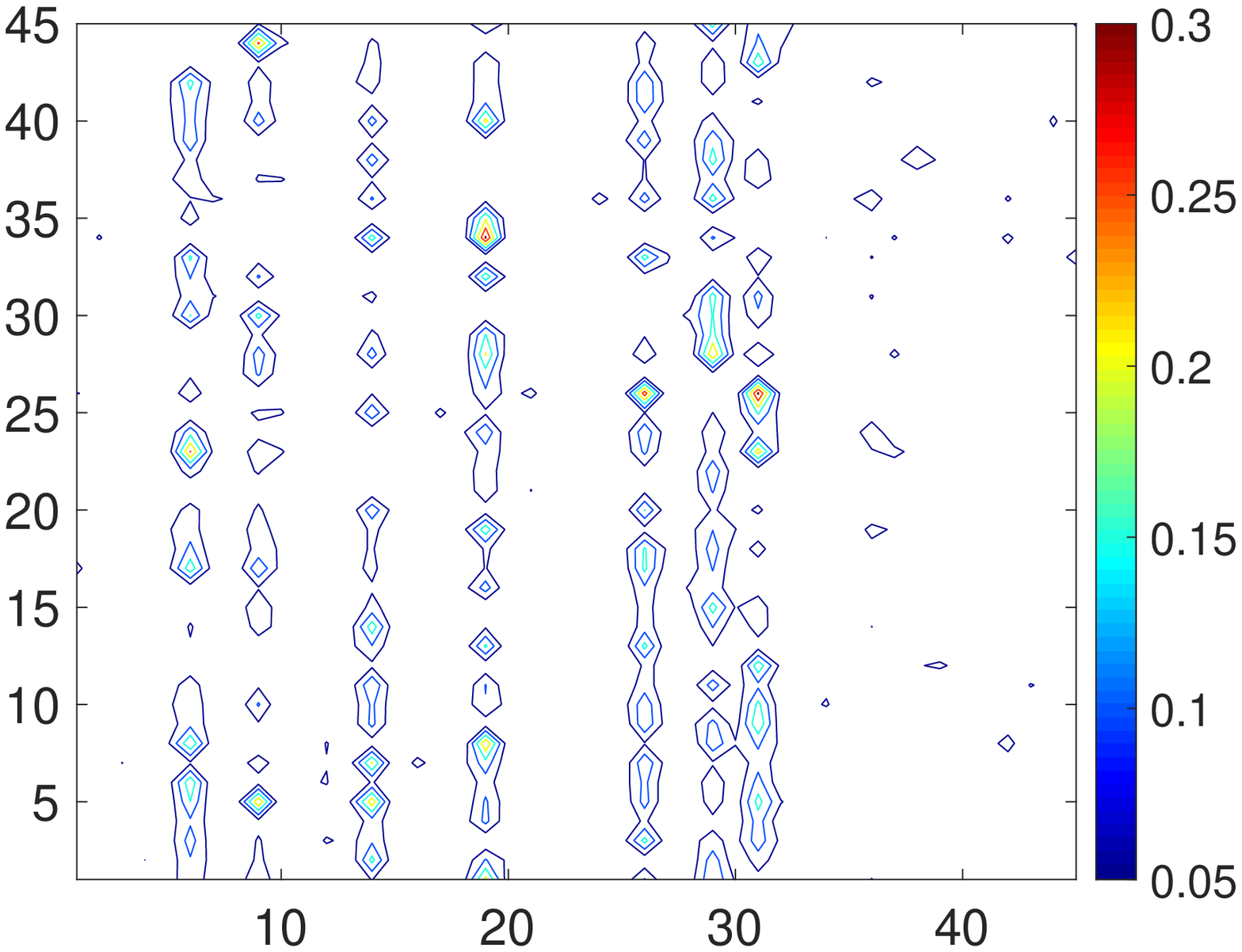}
     \caption{\textit{hypernym}}
     \label{fig:hyper}
    \end{subfigure} 
    \begin{subfigure}{0.27 \textwidth}
    \includegraphics[width=\textwidth]{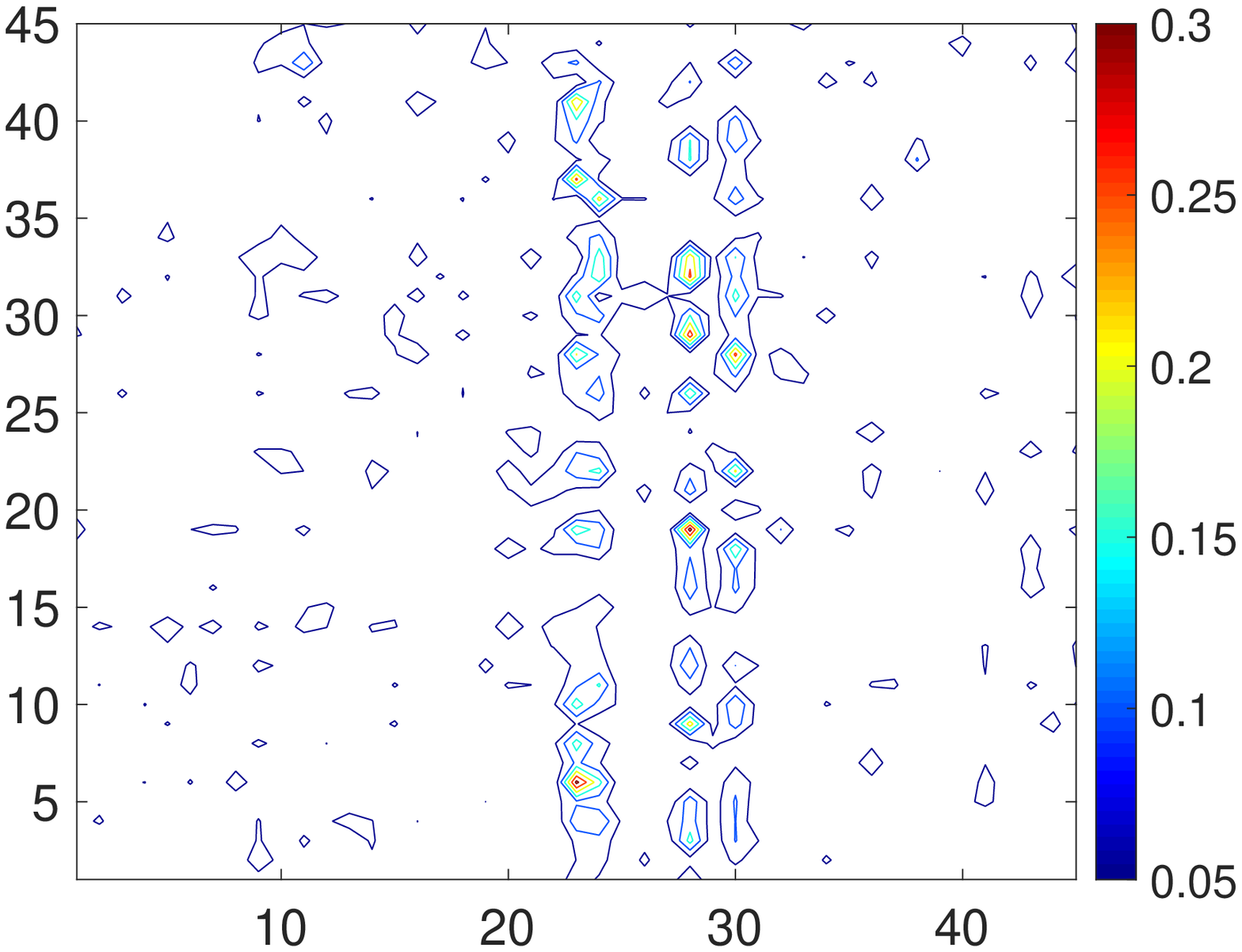}
    \caption{\textit{hyponym}}
    \label{fig:hypon}
    \end{subfigure} 
}

 \mbox{\hspace{-0.3in}
    \begin{subfigure}{0.27 \textwidth}
    \centering
    \includegraphics[width=\linewidth]{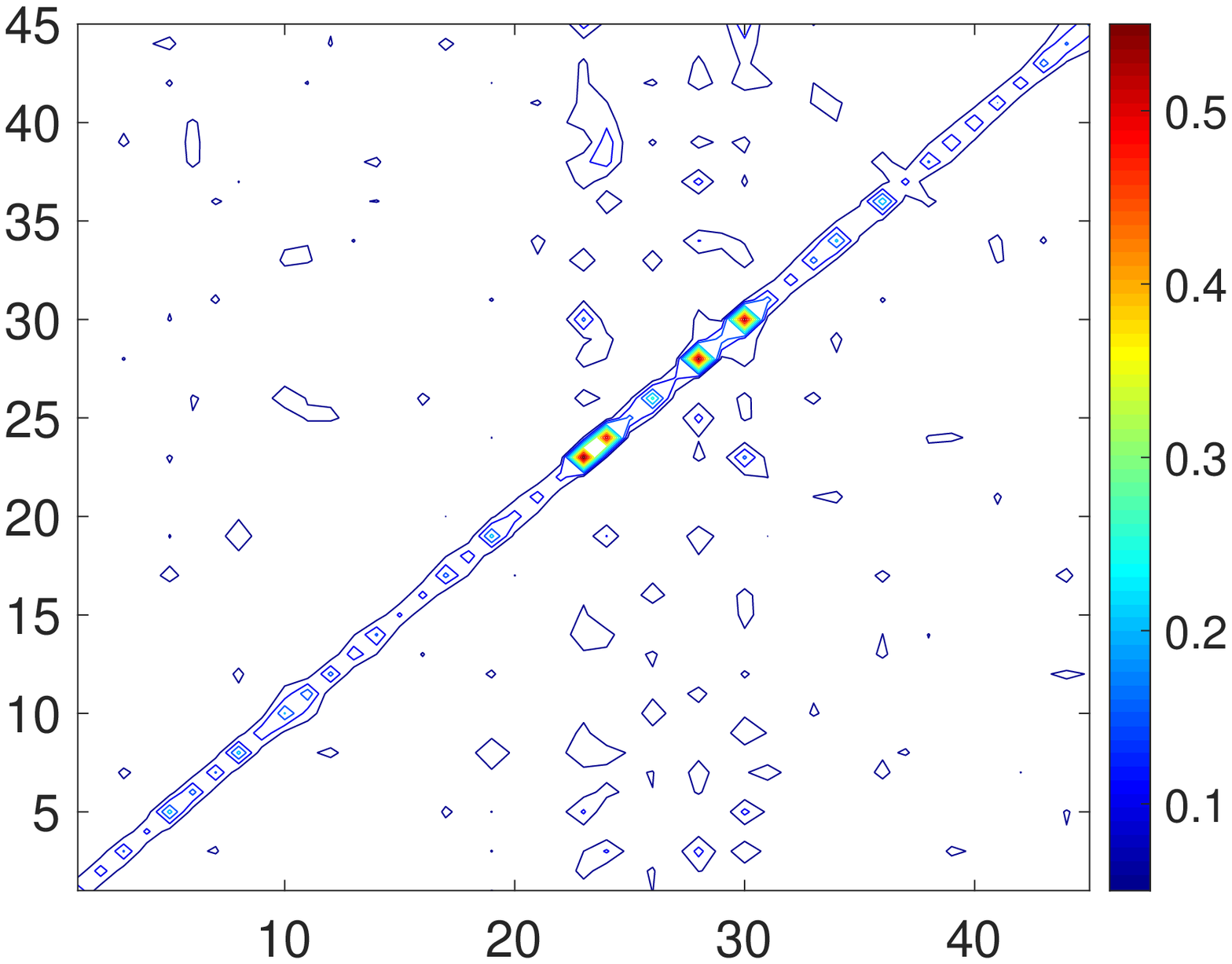}
    \caption{\textit{hypernym} $\otimes$ \textit{hyponym}}
    \label{fig:hyper-hypon}
    \end{subfigure} %
    \begin{subfigure}{0.27 \textwidth}
    \includegraphics[width=\linewidth]{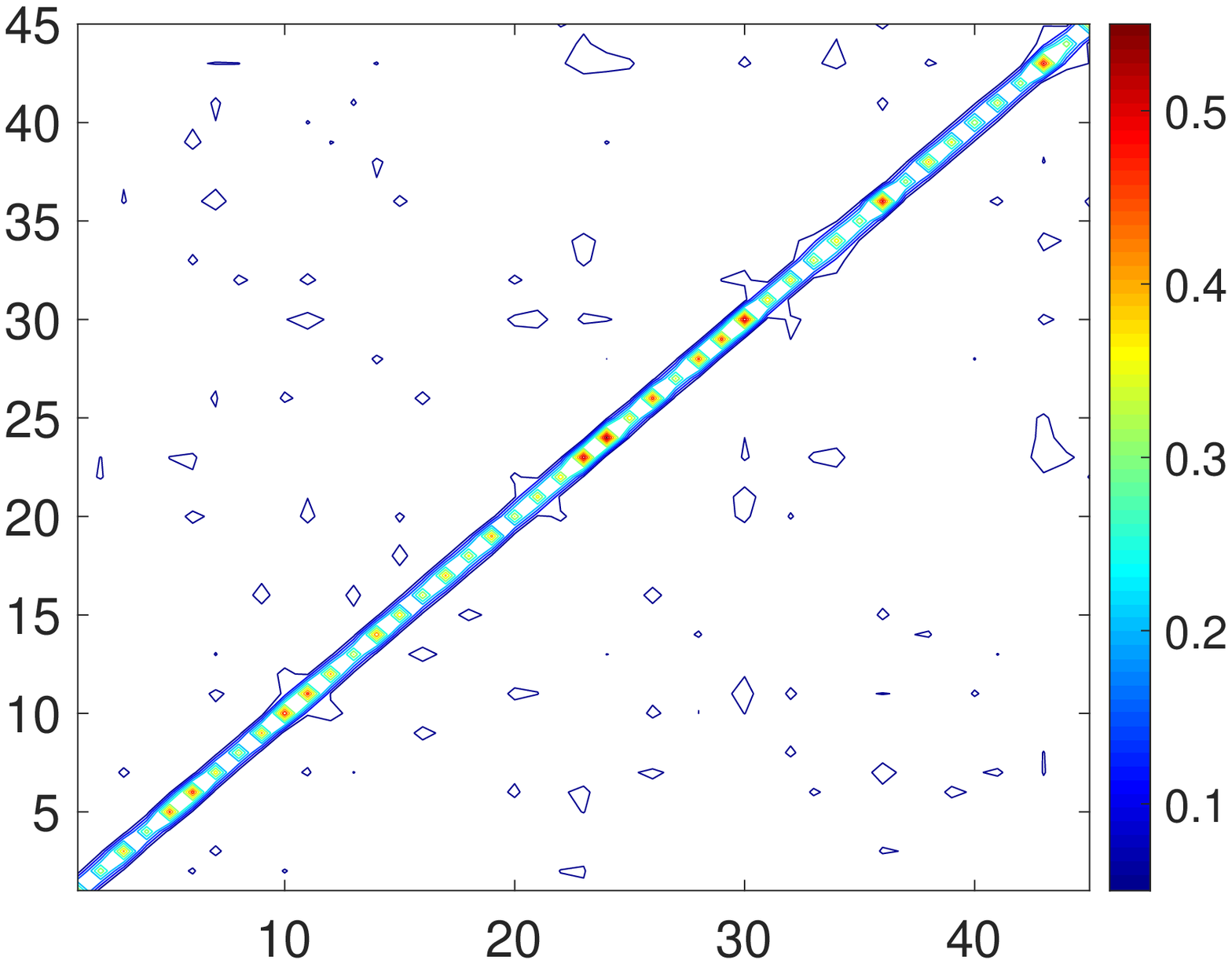}
    \caption{\textit{has\_part} $\otimes$ \textit{part\_of}}
    \label{fig:part-part-of}
    \end{subfigure} %
    \begin{subfigure}{0.27 \textwidth}
    \includegraphics[width=\linewidth]{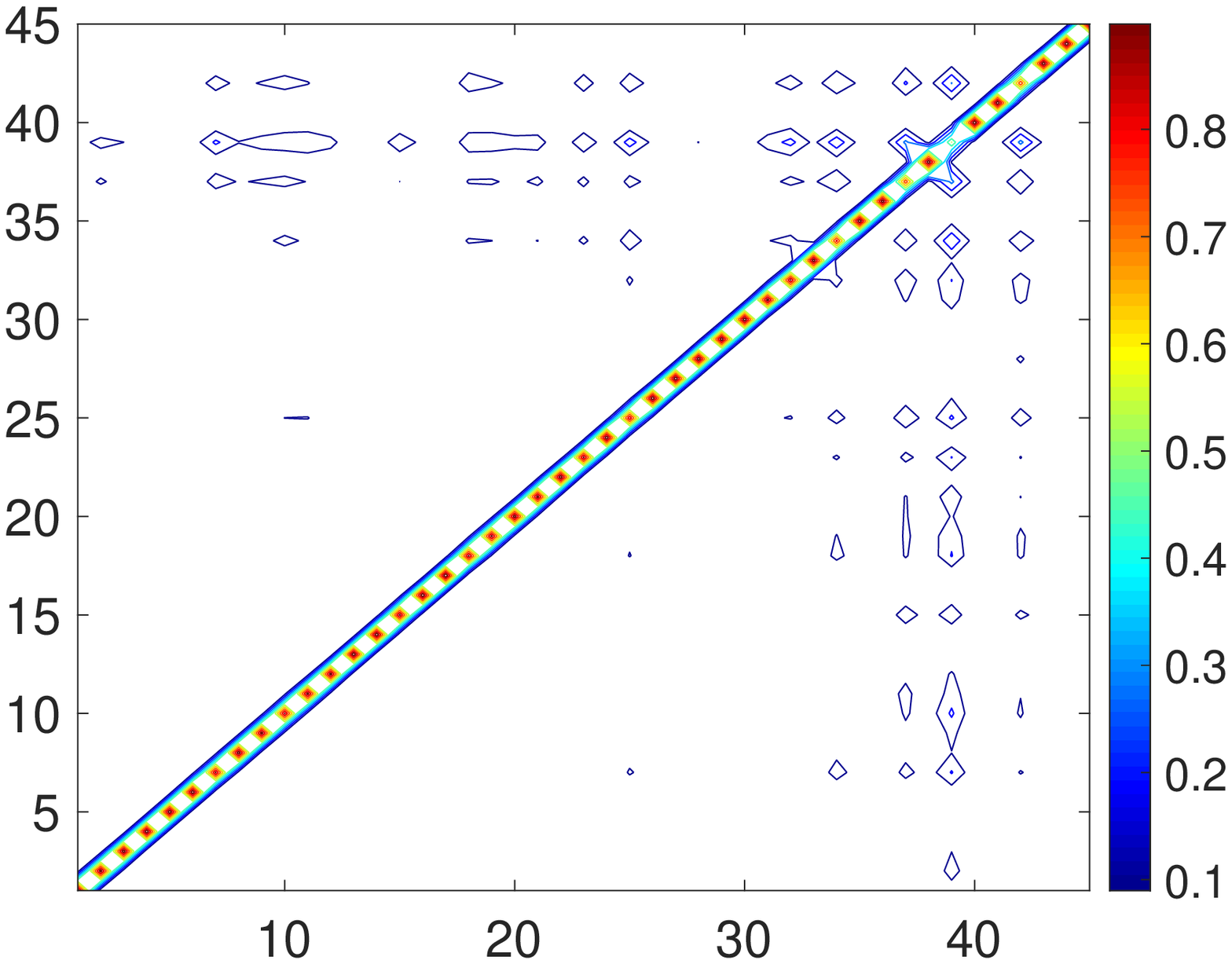}
    \caption{$\widehat{\textit{for}}_2 \otimes \textit{for}_2$}
    \label{fig:for2-iden}
    \end{subfigure} %
    \begin{subfigure}{0.27 \textwidth}
    \includegraphics[width = \linewidth]{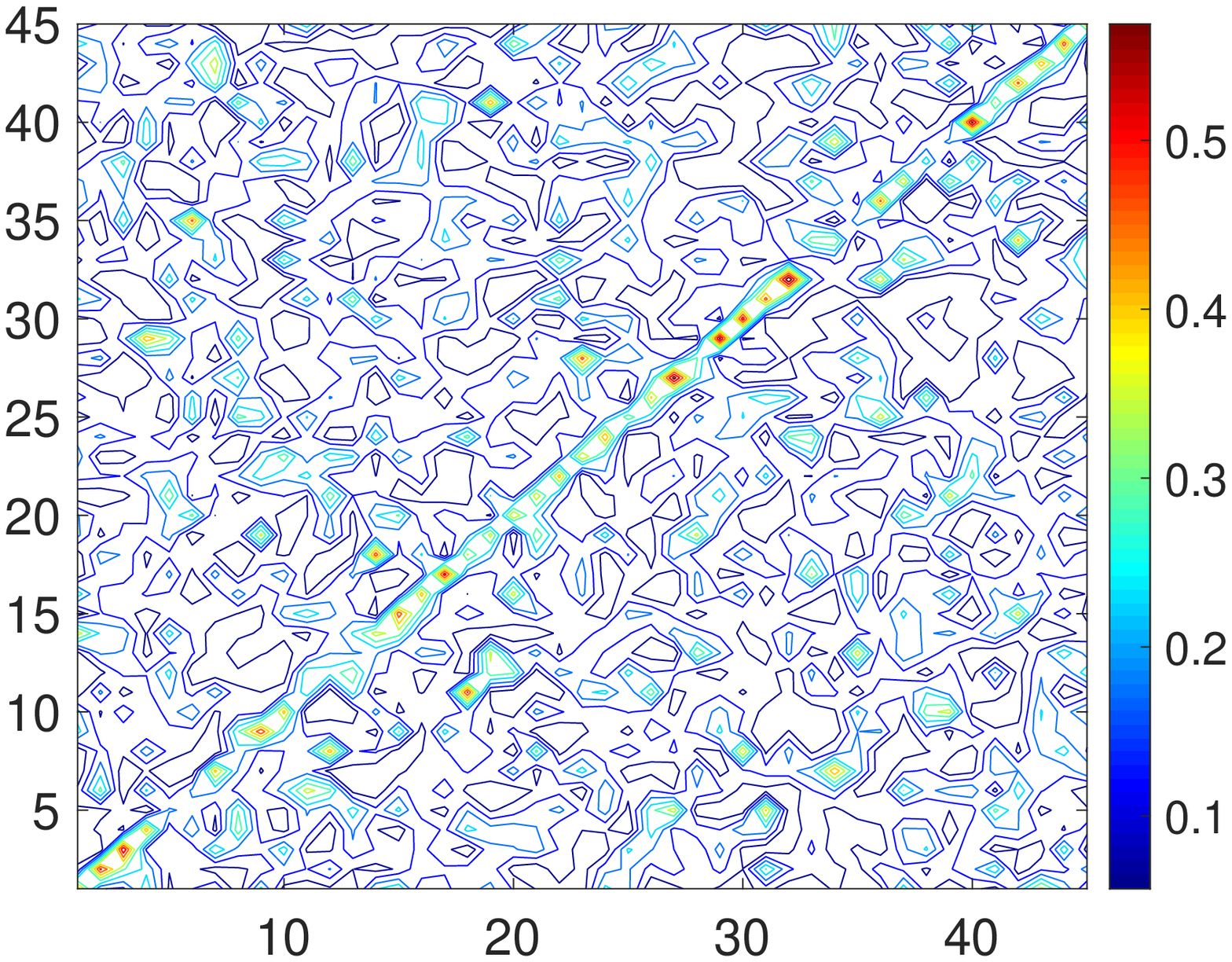}
    \caption{$\widehat{\textit{for}}_2 \otimes$ \textit{winner} $\otimes$ \textit{for}$_1$}
    \label{fig:for2-comp}
    \end{subfigure}
  }
  
  \vspace{0.1in}
  
     \caption{Contours of relation matrices. $\textit{for}_2$, $\textit{winner}$, $\textit{for}_1$ represent the relation \textit{\small award\_nominations./nominated\_for}, \textit{\small winners./award\_winner}, \textit{\small nominees./nominated\_for}, respectively. Figure (\ref{fig:similar},~\ref{fig:verb}): $|\bm{R}^2|$ of symmetric relations; (\ref{fig:hyper},~\ref{fig:hypon}):  $|\bm{R}^2|$ of skew-symmetric relations; (\ref{fig:hyper-hypon},~\ref{fig:part-part-of}): $|\bm{R_1}\bm{R_2}|$ for two inversion relations $r_1$ and $r_2$; (\ref{fig:for2-iden}): $|\hat{\bm{R}} \bm{R}|$ of the relation $\textit{for}_2$; (\ref{fig:for2-comp}): $|\hat{\bm{R}}_3 \bm{R_1} \bm{R_2}|, r_3 = r_1 \otimes r_2$. Matrices  are normalized by their $L_2$ norms for visualization.}
      \label{fig:symmetry}\vspace{0.2in}
 \end{figure}

\subsection{Case Studies on Relation Patterns}

\vspace{0.1in}

The relation matrix representation of SpaceE provides insights into the property of the relation.

\vspace{0.1in}

\textbf{Symmetry/skew-symmetry.} We plot the contours of $|\bm{R}^2|$ for a relation $r$. Figure~\ref{fig:similar} and Figure~\ref{fig:verb} show that for symmetric relations \textit{similar\_to} and \textit{verb\_group} in the WN18 dataset, the matrix $|\bm{R}^2|$ is similar to the identity matrix. Figure~\ref{fig:hyper} and Figure~\ref{fig:hypon} show that  for skew-symmetric relations \textit{hypernym} and \textit{hyponym}, the matrix looks different from the identity matrix. 

\vspace{0.1in}
\textbf{Inversion.} We plot the contours of $|\bm{R}_1 \bm{R}_2|$ for two relations $r_1$ and $r_2$ that are the inversions of each other. As shown in Figure~\ref{fig:hyper-hypon} and Figure~\ref{fig:part-part-of}, $|\bm{R}_1 \bm{R}_2|$ approximates the identity matrix for inversion relation pairs \textit{hypernym} $\otimes$ \textit{hyponym} and \textit{has\_part} $\otimes$ \textit{part\_of} in the WN18 dataset. It confirms that SpaceE is able to model the inversion of the relation by the inversion of the relation matrix.

\vspace{0.1in}
\textbf{Composition.} For three relations $\textit{for}_2$, $\textit{winner}$, $\textit{for}_1$ in FB15k-237 dataset, $\textit{for}_2 = \textit{winner} \otimes \textit{for}_1$, we denote their relation matrices as $\bm{R}_3, \bm{R}_1, \bm{R}_2$. Let $\widehat{\textit{for}}_2$ be the augmented inverse relation of $\textit{for}_2$ and $\hat{\bm{R}}_3$ be its relation matrix. We plot the contours of $|\hat{\bm{R}}_3\bm{R}_3|$ and $|\hat{\bm{R}}_3 \bm{R}_1 \bm{R}_2|$ in Figure~\ref{fig:for2-iden} and Figure~\ref{fig:for2-comp}, respectively. It shows that the matrices $|\hat{\bm{R}}_3\bm{R}_3|$ and $|\hat{\bm{R}}_3 \bm{R}_1 \bm{R}_2|$ look similar to the identity matrix. The values in the diagonal positions are larger than the values in other positions. Thus, $\bm{R}_3 \approx  \bm{R}_1 \bm{R}_2$.

\newpage

\section{Conclusion}
The general philosophy of representation learning is to automatically discover the representations of input data in order to make appropriate decisions.  Two fundamental problems in representation learning  are (i) the form of representations and (ii) the decision function. The mathematical properties of the representation form and the decision function determine the learning ability of the learning machine. 

\vspace{0.1in}
\noindent In this paper, we study the ability of representation learning algorithms in knowledge graph reasoning. We show that the 
relationships in knowledge graphs vary in their logical properties, including injective vs. non-injective, Abelian vs. Non-Abelian, etc. Furthermore,  the reasoning procedure in the knowledge graph builds logical connections between relations, such as inversion and composition. These logical properties and connections exert mathematical constraints on the representation form and the decision function. Improper design of the representation form and the decision function may fail in implementing those logical properties and connections. We demonstrate that the theoretical failures indeed happened in existing methods.

\vspace{0.1in}
\noindent As the solution to implementing all these logical properties and connections, we propose a translation distance-based knowledge graph embedding method called SpaceE using the idea of modeling relations as linear transformations in the entity space. We theoretically demonstrate the ability of SpaceE to model various relation properties, including injective, non-injective, symmetry, skew-symmetry, inversion, Abelian composition, and non-Abelian composition. Qualitative case studies show that the property of learned relation matrices can reflect the symmetry, skew-symmetry, inversion, and composition of relations. Experiments on five benchmark datasets show that our model  obtains competitive results on all benchmarks. On two datasets, FB15k-237 and Yago3-10, which contain many non-injective relations, SpaceE substantially outperforms previous translation distance-based KGE methods, especially on highly non-injective relation triples.

\vspace{0.1in}
\noindent Although our method in this paper is proposed and tested only in knowledge graph reasoning, our design can potentially be applied to other fields that involve complex logical properties and connections, such as many problems in causal discovery. It is a long road to find a general representation learning solution for all kinds of logical relations. It is admitted that we have only made a small step forward down the road.

\vspace{0.1in}

\vspace{0.5in}

\bibliographystyle{plainnat}
\bibliography{kge}
\end{document}